\documentclass{article}

\usepackage{amsmath, amsthm, amssymb, amsfonts, mathtools, mathrsfs}
\usepackage{thmtools}
\usepackage{graphicx}
\usepackage{geometry}
\usepackage[authoryear]{natbib}
\usepackage{enumitem}
\usepackage{algorithm}
\usepackage{algpseudocode}
\usepackage{caption}
\usepackage{subcaption}
\usepackage{comment}
\usepackage[dvipsnames]{xcolor}
\usepackage{hyperref}
\usepackage{wrapfig}
\usepackage{authblk}

\newtheorem{assumption}{Assumption}
\newtheorem{condition}{Condition}
\newtheorem{remark}{Remark}
\theoremstyle{definition}
\newtheorem{example}{Example}
\begin{document}

\title{Structured Output Regularization: a framework for few-shot transfer learning}

\author{%
  Nicolas Ewen\thanks{Corresponding author. Email: gnic@my.yorku.ca},  Jairo Diaz-Rodriguez, Kelly~Ramsay\\
  
 \vspace{0.5em}
 
  Department of Mathematics and Statistics\\
  York University\\
  Toronto, Ontario M3J 1P3
  
}

\maketitle

%

%





\begin{abstract}
Traditional transfer learning typically reuses large pre-trained networks by freezing some of their weights and adding task-specific layers. While this approach is computationally efficient, it limits the model's ability to adapt to domain-specific features and can still lead to overfitting with very limited data. To address these limitations, we propose Structured Output Regularization (SOR), a simple yet effective framework that freezes the internal network structures (e.g., convolutional filters) while using a combination of group lasso and $L_1$ penalties. This framework tailors the model to specific data with minimal additional parameters and is easily applicable to various network components, such as convolutional filters or various blocks in neural networks enabling broad applicability for transfer learning tasks. We evaluate SOR on three few shot medical imaging classification tasks and we achieve competitive results using DenseNet121, and EfficientNetB4 bases compared to established benchmarks.

\end{abstract}

\section{Introduction}
Transfer learning is often used in deep learning when  data is limited, such as in medical imaging applications \citep{kim2022transfer}. 
Foundation models, that is large, publicly available, pre-trained models, are often fine-tuned for such tasks where little data is available \citep{wang2023real,zhang2024challenges,khan2025comprehensive}. Beyond freezing 
part of a model to reduce overfitting, 
various techniques can increase training data 
such as data augmentation,
and self supervised learning.
These methods can reduce overfitting \citep{chollet2021deep, wang2023real, ewen2021targeted}, but still struggle when there is little data available \citep{wang2023real}. 

\begin{figure}[t!]
     \centering
         \centering
         \includegraphics[width=0.55\textwidth]{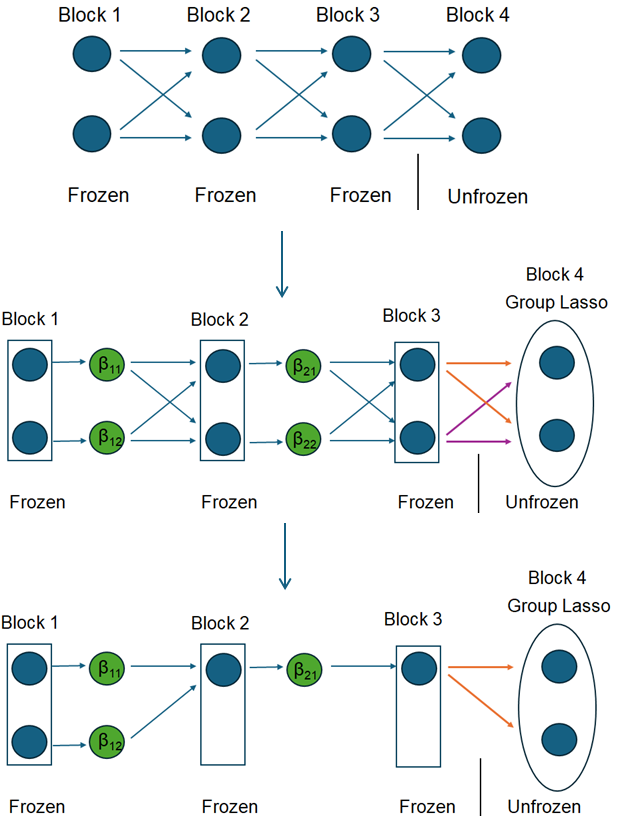}
         \caption{Outline of SOR. We start with a mostly frozen model (top), then add new weights $\beta$ and penalties (middle), resulting in a regularized model (bottom).}
     \label{combinedIntro}
\end{figure}

We propose a new approach, \emph{Structured Output Regularization (SOR)}, a simple framework that adapts and prunes pretrained networks using very little labeled data. Instead of unfreezing internal weights, SOR keeps internal structures frozen, e.g., convolutional filters or higher-level blocks, and regularizes their \emph{outputs}. Specifically, we freeze  internal structure weights, we add new weights between each frozen structure, penalized via lasso penalty to encourage sparsity, and train the network. Structures whose new weights are driven to zero can be removed, yielding a smaller, task-tailored model without training the full parameter set. To regularize the final layer structures, SOR applies group lasso. see Figure \ref{combinedIntro}.

SOR also has the benefit of structured pruning, 
which 
reduces the resources needed for a model, while maintaining performance \citep{he2023structured,li2023model}. 
Many existing pruning methods either rely on 
arbitrary heuristics, and thus need to be re-trained after pruning \citep{he2023structured,li2023model,zhang2022fpfs}, 
or require the full model to be trained from scratch, which is difficult when data is scarce \citep{chen2021only,ochiai2017automatic}. 
This limits methods' usefulness when data is limited, as otherwise they could be used to shrink a model during training, and thus help to reduce overfitting.
SOR avoids both drawbacks: it does not need retraining, and can be applied in 
scenarios with little data. 

We demonstrate SOR on convolutional layers, DenseNet blocks, and EfficientNet blocks, but the framework applies more generally to models composed of sequential subfunctions as well. Across three few-shot medical imaging classification tasks, SOR achieves competitive performance while mitigating overfitting, using DenseNet-121 and EfficientNet-B4 models, relative to established transfer learning baselines \citep{wang2023real}
, see Section \ref{experiments}.

\paragraph{Summary of Contributions.}
\begin{itemize}
    \item We introduce \emph{Structured Output Regularization (SOR)}, a data-efficient transfer learning framework that keeps pretrained models frozen and applies lasso / group-lasso penalties to structure-level outputs, reducing overfitting and allowing pruning with few additional parameters.
    \item We provide implementations for convolutional layers, DenseNet-121 blocks, and EfficientNet-B4 blocks; SOR requires no additional pretraining and integrates with standard transfer learning pipelines.
    \item Across three few-shot medical imaging classification tasks, SOR often outperforms few shot  baselines, in many cases achieving better results using less data than the benchmarks, see Section~\ref{experiments}.
\end{itemize}

    

\subsection{Related work}
\textbf{Transfer learning.}
Transfer learning generally refers to adapting a trained model to perform a different task. The new different task is often called a downstream task.
Transfer learning from large pre-trained models to downstream tasks with little data is an area of open research. 
When data is scarce, full fine-tuning, (i.e. training all parameters in the whole model), of the pre-trained model can quickly overfit \citep{chollet2021deep,wang2023real,han2024facing,jia2022visual}. To deal with this, various methods such as self supervised learning can be used to further pre-train using more data relevant to the task, however, a final fine-tuning stage is still required \citep{wang2023real, ewen2021targeted}. Reducing the total number of parameters tuned during training is often done to avoid overfitting \citep{chollet2021deep,han2024facing,jia2022visual,wang2023real}. The most common method of reducing the number of parameters is to perform a partial fine-tuning of the model \citep{chollet2021deep,han2024facing,jia2022visual,wang2023real}. One drawback of this is that large parts of the pre-trained model will not get adapted to the new task or domain. A recent advance in vision transformers has been to freeze the model, add extra trainable parameters to the model input, and to only train those new parameters \citep{han2024facing,jia2022visual}. However, this adds to the complexity of the model, reducing it's interpretability, and still adds in a large number of new parameters in many models. There is a need for a transfer learning method that greatly reduces the number of parameters, while allowing all parts of the model to adapt to the downstream task. 

\textbf{Structured pruning.}
Structured pruning is a common technique to reduce the size of a model by removing groups of parameters that form structures within a model \citep{he2023structured,bai2021explainable,li2023model}. Some authors use heuristics to remove structures\citep{zhang2022fpfs}, which can reduce model performance, resulting in the need to re-train.
Some authors add a simple penalty to the loss function, such as an $L_1$ on the model weights, and decide on cut-offs
 \citep{kumar2021pruning}. 
Various methods for pruning exist that prune while training the entire network. 
Network slimming 
 requires both the the full model to be trained and iterative re-training \citep{liu2017learning}.
Some authors propose training the model with additional ``masking'' or ``gate'' parameters that help set some structures in the CNN to zero while leaving the others unchanged \citep{ding2021resrep, he2023structured}. 
A number of authors use a group lasso penalty during training to eliminate certain structures in the model \citep{wen2016learning, chen2021only, ochiai2017automatic}. OTO \citep{chen2021only} penalizes specific zero-invariant groups. 
To our knowledge, there are no methods that do not require re-training, prune the model using available data, and leave many parameters frozen.

\section{Background}

\subsection{Convolutional neural networks}
We will use convolutional neural networks (CNNs) in this paper as examples, and now provide a brief overview of them.
CNNs are neural networks with at least one convolutional layer. A convolutional layer typically consists of several subfunctions called filters, which perform identical operations, but use different weights. Each filter has its own unique set of parameters, and produces a part of the layer's output. The filters in a layer do not interact with each other. Each filter in a layer has the same number of parameters, which depends on the kernel size, and the number of input channels. For example, a filter that operates on an RGB image, with a 3x3 kernel would have 9 parameters for each input channel, one for each of R, G, and B, and one bias value for a total of 28 parameters. For 2-dimensional (2D) images, a convolutional layer will take an image with multiple 2D matrices called channels as input, and output a derived 2D image with one channel for each filter in the layer. The input and output image can have different number of channels, height and width. 
For further reading see \cite{chollet2021deep},  \cite{goodfellow2016deep} and \cite{ brownlee2019deep}.

\subsection{Problem}

We first introduce notation to formalize the problem. We are concerned with \textit{transfer learning}. We assume there is some machine learning ``task'' for which we wish to modify an existing model ($M$) to complete. In particular, we suppose we have $n$ pairs of input-output task-specific data $(X,V) = \{x_i, y_i\}_{i=1}^{n}$, which is related to a much larger, more general dataset $(X',V') = \{x'_i, y'_i\}_{i=1}^{N}$, where $N >> n$. It may be the case that $(X,V) \subset (X',V')$. We also suppose we have access to a model $M$ and the model has been pre-trained on the larger dataset. The model $M\coloneqq M(\theta)$, where $\theta\in\mathbb{R}^p$, $p >> n$, and has $L$ sequential blocks. That is, it can be expressed as a composition of functions $f_L \circ \ldots \circ f_2 \circ f_1$. Each block or subfunction $f_i$ can be composed into multiple layers, and produces a sequence of at least two outputs $a_{ij}, j=1,\ldots,J_i, J_i \geq 2$. The aim is to adjust and shrink $M$ to do the task, using $(X,V)$. We aim to do this in such a way that avoids arbitrary heuristics, iterative fine-tuning, and extreme overfitting. 
We will approach this problem from two directions: pruning parts of $M$ to reduce the total number of parameters and freezing parts of $M$ to reduce the number of parameters that are being trained.
%
A common example of such a set-up can be found in medical imaging classification tasks where data is limited. For instance, we have a convolutional neural network base $M$ pre-trained on ImageNet. 
Here, ImageNet is the large dataset without the task specific data i.e., $(X',V') / (X,V)$. 
The goal is to adjust $M$ given the new medical imaging data $(X,V)$. 

\begin{wrapfigure}{r}{0.4\textwidth}
     \centering
         \centering
         \includegraphics[width=0.35\textwidth]{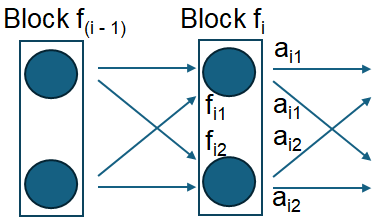}
         \caption{An example of the notation showing a block, $f_i$, a subfunction of a block, $f_{i1}$, and it's output $a_{i1}$. }
     \label{notation}
\end{wrapfigure}

\subsection{Preliminary notation}
Before introducing SOR, we introduce some notation. 
Let the input to block $f_i$ be denoted as $Z_i$. Let the subfunction, or structure, of $f_i$ that produces output $a_{ij}$ be denoted as $f_{ij}$. Let $W$ be the set of parameters of the model, let $W_i$ be the subset of parameters $W$ that are in block $i$, and let $w_i$ be the subset of $W_i$ that interacts with the inputs to block $i$. Let the subset of $W_i$ that is used by $f_{ij}$ be denoted as $W_{ij}$.
Each block can be expressed as 
\begin{equation*}
\begin{split}
   f_i(Z_i, W_i) &= [f_{i1}(Z_i, W_{i1}),  \ldots, f_{iJ_i}(Z_i, W_{iJ_i})] \\
   &= [a_{i1}, a_{i2}, \ldots, a_{iJ_i}] = Z_{i+1}.
   \end{split}
\end{equation*}


Observe that within each block $f_i$, there are $J_i$ sub-functions, $f_{ij}, j = 1,\ldots,J_i$, that each produce one channel of that block's output $a_{ij}$. 
Furthermore, each sub-function is associated with a portion, $W_{ij}$, of the block's trainable parameters, $W_i$.

\subsection{Assumptions and Conditions}
We now introduce some assumptions on the pretrained model $M$ that are necessary for SOR. 
Let $w_{i,k}$ be the subset of $w_i$ that interacts with input $a_{(i-1),k}$.  
\begin{assumption}\label{ass2}
    Assume that for each block, $f_i$ in the model, that if $\forall u \in w_{ik}, u=0$ then changing $Z_i$ to $Z_i'$ by setting $a_{(i-1),k} = 0$ results in $f_L \circ \cdots \circ f_i(Z_i,W_i) = f_L \circ \cdots \circ f_i(Z_i',W_i)$.
\end{assumption}
Assumption \ref{ass2} says that for each block in the model, if all parameters in the block that interact with a part of the input are zero, then setting that part of the input to zero does not change the model's output. An example where this would not hold is if a block passed it's input unchanged as part of it's output. If blocks are carefully defined, this assumption holds for most neural networks. 
It is helpful to compare our Assumption \ref{ass2} to the following condition, which is commonly imposed, either explicitly or implicitly, in the literature \citep{chen2021only,ma2022phase,he2023structured,ochiai2017automatic}
\begin{condition}\label{cond1}
    That for each block, $f_i$ in the model, that if $\forall u \in W_{ij}, u=0$ then $a_{(i),j} = 0$.
\end{condition}
Most previous works assume Condition \ref{cond1} holds. 
On the other hand, we only require the weaker Assumption~\ref{ass2}, which is a result of our group lasso grouping strategy. For more details, see Remark~\ref{rem1}.
\begin{assumption}\label{ass4}
    Assume that for each block, $f_i$ in the model, that if  $a_{(i-1),j} = 0$ then changing $Z_i$ to $Z_i''$ by removing $a_{(i-1),j}$ results in $f_i(Z_i,W_i) = f_i(Z_i'',W_i)$.
\end{assumption}
Assumption \ref{ass4} says that if a particular output $a_{(i-1),j}$ from a block is always zero, removing that output, along with parameters unique to $f_{(i-1),j}$ does not change the next block's behavior. This assumption allows us to prune the model, and an example where it may not hold is if a block uses the number of inputs it receives. This assumption holds for most neural networks.
Assumption \ref{ass4} is commonly imposed, sometimes implicitly, in prior works on model pruning for neural networks and CNNs \citep{ma2022phase,li2023model,kumar2021pruning, ochiai2017automatic, chen2021only}. 
\section{Structured Output Regularization}\label{SOR}
We now present structured output regularization (SOR), our proposed framework for simultaneous transfer learning and pruning of pre-trained networks when little data is available. 
At a high level, given a pre-trained model, SOR adjusts and prunes the model without requiring further fine-tuning. Specifically, new weights are added to subvectors of the parameter vector $W$, called structures, which the analyst would like to regularize. The model is then trained with the available, task specific data. A combination of group lasso and $L_1$ penalties are applied where appropriate, giving a regularized model. Structures whose outputs have been set to zero by either the new weight, or the group lasso can then be pruned, which gives a reduced model. 
This is outlined in Algorithm \ref{alg:overview}. 
Let us now be more precise.

\begin{algorithm}
\caption{Outline of SOR}\label{alg:overview}
\begin{algorithmic}[1]
\State Input: A full pre-trained model, $M$.
\State Define blocks $f_i$
\State Choose blocks, $f_i$, to freeze.
\State Add new weights, $\beta_i$, to outputs, $a_i$, between frozen structures, $f_i$.
\State Apply $L_1$ and group lasso penalties to $\beta$'s and unfrozen blocks respectively. 
\State Train model $\tilde{M}$ with penalties.
\State Prune structures with zeroed outputs: $(\beta_{ij}=0, (w_{ik}|\forall u \in w_{ik}, u=0))$.
\State Output: A regularized model.
\end{algorithmic}
\end{algorithm}

\textbf{Step 1: Define blocks.} The first step is to define the blocks, $f_i$, which are sequential subfunctions of the model $M$, whose outputs, $a_i$, the researcher may want to regularize. 
Blocks, $f_i$, should be chosen in such a way that they do not violate the assumptions. 

For example, suppose our model is a CNN with a single dense layer, $z$, and a convolutional base that consists of three convolutional layers. Then $W_B$ is the set of trainable parameters in the convolutional base, $h$, $x$ is the input vector for the dense layer, $\sigma$ is some activation function, and $w_L$ and $b$ are the usual weights and bias for a dense layer. $\tilde{G}$ represents a flattening function that vectorizes the output of the convolutional base.
We then have ${M} = z(\tilde{G}({h}(X, W_B)), w_L, b)$, with a convolutional base, $h(X, W_B)$, and a single layer dense top, $z(x, w_L, b) = \sigma(w_L^T \cdot x + b )$.
In this example, the researcher may decide that they would like to regularize the outputs of each convolutional layer, and may be interested in removing unnecessary filters from each of the layers.
The researcher could then choose to define each convolutional layer as its own block $f_i$.

\textbf{Step 2: Freeze blocks.} The next step in SOR is to choose the subset of blocks, i.e., $f_i$, to freeze  during training. This is done by selecting $i_c < L$, and freezing all parameters $W_i$ in for every block $i\leq i_c$, leaving parameters for blocks $i > i_c$ unfrozen.
This means all blocks up to a point are frozen, not including the last block, and all subsequent blocks are not frozen, meaning that they will be retrained on the task-specific data. This is common in many transfer learning strategies \citep{chollet2021deep}. See Figure \ref{combinedIntro} top for an example.
This step is necessary to avoid heavy overfitting.
Instead of training and penalizing all parameters in a model, $M$, we allow all the parameters in some blocks, $f_i$, to be frozen. This greatly reduces the total number of parameters in the model to be estimated during retraining, which helps reduce overfitting when data is limited. 
The value $i_c$ can be decided based on how much available training data there is, but $i_c=L-1$ is a general standard.

Continuing our example, 
we assume that the researcher decides that the amount of training data available, $X$, is enough to train the final dense layer, and a few more parameters, but not enough to train another layer as well. 
The researcher then chooses to freeze all three convolutional layers to avoid heavily overfitting. This leaves the researcher with $f_1$, $f_2$, and $f_3$ being frozen convolutional layers, and $f_4$ being an unfrozen dense layer. See Figure \ref{combinedIntro} top.
For further examples of this, see Section~\ref{paper ex} in the supplementary materials where we apply our method to specific models. 

\textbf{Step 3: Add new weights between frozen blocks.} 
The next step is to insert a  new set of parameters $\beta$ between frozen blocks. As $M$ is sequential, by regularizing the output of each block, we are equivalently regularizing the input of each subsequent block. 
Let $\beta_i$ be the subset of new parameters ($\beta$) that are applied to the output of block $i$ (or equivalently to the input of block $i+1$). 
Applying new weights $\beta_i$ to the outputs of certain blocks $f_i$ allows for the model to be adjusted and pruned with fewer parameters. 
Equivalently, we can say $\beta_{i-1}$ is applied to the input of block $i$, denoted $f_i$, when the preceding block $f_{i-1}$, as well as block $f_i$ are frozen. 
To be more precise, each $\beta_i$ consists of $J_i$ scalar values. Each $f_i$, $i \leq i_c$, except for the last one, $f_{i_c}$, will become an $\tilde{f}_i$ :
\begin{equation*}
\begin{split}
    \tilde{f}_i(Z_i, W_i, \beta_i) = [\beta_{i1} \times f_{i1}(Z_i, W_{i1}),  \ldots,\\ \beta_{iJ_i} \times f_{iJ_i}(Z_i, W_{iJ_i})],
    \end{split}
\end{equation*}
where $\beta_{ij}$  multiplies with the output of sub-function $f_{ij}$. 
In our earlier example, all blocks up to and including $f_3$ are frozen, therefore $c=3$. The researcher will insert a $\beta_i$ after $f_1$ and $f_2$, but not after $f_3$, instead putting a group lasso penalty on the parameters in the dense layer to regularize the outputs of $f_3$. This would mean that the researcher would put $\beta_1$ after $f_1$, $\beta_2$ after $f_2$, but the researcher would not put a $\beta_3$ after $f_3$. Instead, they would put a group lasso penalty on $f_4$ grouped by input to regularize $a_3$. See Figure \ref{combinedIntro} middle.

\textbf{Step 4: Regularization} 
Next, we enforce sparsity in the blocks. Between frozen blocks we apply a lasso penalty to the new weights $\beta$. This strategy effectively reduces the number of parameters by selecting structure via lasso. Additionally we apply a group lasso regularization to inputs of the first unfrozen layer. 
Using a combination of group lasso and new parameters, rather than just new parameters, is necessary to avoid over parameterization, see Remark~\ref{OverParaRem}.


The reasoning for our example is as follows: The researcher wishes to regularize the outputs of $f_1$, $f_2$, and $f_3$. The output of $f_3$ can be regularized with a group lasso penalty on the currently unfrozen parameters in $f_4$, so the researcher will implement such a penalty for the output of $f_3$. For the outputs of $f_1$ and $f_2$, they become the inputs for $f_2$ and $f_3$ respectively, which are both frozen. Therefore, we cannot put a penalty on the frozen parameters of $f_2$ and $f_3$ to regularize their inputs, and must instead use a $\beta_1$ and $\beta_2$ to regularize $a_1$ and $a_2$ respectively.

%



\begin{remark}\label{OverParaRem}
The goal is to regularize and eliminate certain blocks' outputs within the model. For a particular output of interest, $a_{ij}$, if it can be regularized with a group lasso penalty, then there is no reason to apply a new weight $\beta_{ij}$ to also regularize it. Applying a new weight $\beta_{ij}$ to $a_{ij}$ in this case adds an extra parameter to the model that has no purpose. In a situation where there is very limited available task specific data, adding extra unnecessary parameters will lead to heavier overfitting of the model.
\end{remark}

\textbf{Step 5: Train.}
The next step is to train the new model $\tilde{M}$ on the task specific data.
We now introduce the regularization equations. 
Each $\beta_{ij}$ is initialized to 1, so that on the first forward pass the new model $\tilde{M}$ produces the same output as the old model $M$. The equation for the new model, $\tilde{M}$, is as follows:
\begin{equation*}
\begin{split}
    \tilde{M}(X, W, \beta) = f_L(\ldots f_{i_c}( \tilde{f}_{i_c-1}(\ldots\tilde{f}_2(\tilde{f}_1(X, W_1, \beta_1),\\ W_2, \beta_2), \ldots , W_{i_c-1}, \beta_{i_c-1}), W_{i_c}),\ldots,W_L).
    \end{split}
\end{equation*}
Let $\Omega_{0}$ be the set of indices, $1\leq i \leq L$, such that their associated blocks' outputs do not have new weights $\beta_i$, excluding the final block, $L$, and let $\Omega_{new}$ be the set of indices $i$ such that they do have new weights $\beta_i$.
We aim to minimize the following loss function, where $l(\beta, W)$ is a loss function for the specific task, $\lambda_1$ and $\lambda_2$ are the penalty coefficients for the new weights, $\beta$, and the group lasso weights, 
 $u$, respectively:
\begin{multline*}
     \min\limits_{\beta \in \mathbb{R}^n, W \in \mathbb{R}^m} \psi(\beta, W) =  l(\beta, W)   \\+ \lambda_1 \sum_{i \in \Omega_{new}}\sum_{j = 1}^{J_i}|\beta_{ij}| +\lambda_2 \sum_{i\in \Omega_{0}} \sum_{k=1}^{J_i}\sqrt{\sum_{u \in w_{(i+1),k}}u^2}.
\end{multline*}
Following our previous example, we now have a new model, $\tilde{M}$ with new weights $\beta$:
\begin{align*}
    \tilde{M} &= z(\tilde{G}(\tilde{h}(X, W_B, \beta)), w_L, b)\\ 
    & = f_4(f_3(\tilde{f_2}(\tilde{f_1}(X, W_1, \beta_1), W_2, \beta_2), W_3), W_4).
\end{align*}
We aim to minimize the following loss function:
\begin{equation*}
\begin{split}
    &\min\limits_{\beta \in \mathbb{R}^n, w_L \in \mathbb{R}^m, b \in \mathbb{R}^{J_{L}}} \psi(\beta, w_L, b) = \\ &l(\beta, w_L, b) + \lambda_1 \sum_{i=1}^{2}\sum_{j = 1}^{J_i}|\beta_{ij}| + \lambda_2 \sum_{i=1}^{J_{3}}\sqrt{\sum_{j=1}^{J_{4}}w_{L_{ij}}^2}.
    \end{split}
\end{equation*}

\textbf{Step 6: Prune (optional).} The last step is to prune the model if desired and if possible.
After training, we can remove any parameters unique to $f_{ij}$ where $\beta_{ij} = 0$ or any parameters unique to $f_{ik}$ where $\forall u \in w_{ik}, u=0$. This prunes the model without affecting its output.
In our example, if any outputs are removed, the researcher can also remove the filters that produced them see Figure \ref{combinedIntro} bottom.

\begin{remark}\label{rem1}
A number of recent pruning strategies involving group lasso act on the parameters of the structures they are trying to eliminate. This requires Condition \ref{cond1} to be true, that if all parameters in a structure are zero, then the output of that structure is also zero. If we instead use a group lasso penalty on the next block, grouped by the current block's outputs, we get the same result which only requires the weaker Assumption \ref{ass2}. This strategy can also make the setup simpler, as we no longer need to check which parameters contribute to an output, nor check Condition \ref{cond1}. 
For example, we expect that grouping penalties based on inputs, rather than outputs will usually be better for CNNs because Assumption \ref{ass2} tends to be true more often than Condition \ref{cond1}. Other methods for CNNs that rely on Condition \ref{cond1} need to also assume that all activation functions map zero to zero, which may not always be true. However, in most neural networks, Assumption \ref{ass2} is true, as all inputs are usually multiplied by some set of parameters.
\end{remark}
\subsection{Advantages of SOR}
The main advantage of SOR is the ability to adjust a pre-trained model using very few parameters.  Another major advantage of SOR is it's simplicity, both in terms of use, and in implementation. It is simple to use since the model does not need to be iteratively retrained to recover performance. It is simple to implement because it does not require a detailed understanding of the inner workings of the structures in the blocks, such as the convolutional base in our example, and applies to many structures in general. It also has the added benefit that it can perform structured pruning.
SOR reduces some outputs $a_{ij}$ to zero. Since the output is zero, the parameters unique to $f_{ij}$ that created it can be removed without affecting the output of the model. Since the model is unaffected by the pruning, it does not need to be retrained. 

\subsection{Implementation details}
We implemented SOR as a layer that intercepts the data between consecutive blocks. It takes the output from a block and multiplies each channel of it by it's associated new weight. It then passes the changed data to the next block. In this way, we are not concerned about many of the details of what happens in the block, as we only need to focus on the output data.

\section{Experiments} \label{experiments}
We demonstrate our method using two sets of experiments. The first set of experiments are run on a toy dataset ``Noise and Box'' as a proof of concept. The second set of experiments are run on three real-world medical imaging datasets that have been used as benchmarks for few shot learning and transfer learning performance for foundation model adaptation \citep{wang2023real}. Our experiments demonstrate the performance of SOR on two different types of convolutional base blocks, as well as on filter outputs. All experiments were run using Google Colab. 

\subsection{Noise and Box experiments}
The Noise and Box datasets are a toy example as a proof of concept, complete details of the dataset and generation process can be found in the supplementary material. 
The task for this dataset is to classify whether or not there is a `box', or square, in the image. 
We tested three levels of training data (10, 100, and 1000 images), two noise upper bounds (0.1, and 1.0), two different numbers of filters per layer (3, and 10), and three magnitudes of $\lambda$ (0.05, 0.5, and 5). Each experiment was repeated using a different seed 100 times. Each time, a new training set and testing set were generated. We then took the sample mean and sample standard deviation of the training set, and use them to standardize both the train and test sets.

For each iteration, we trained a simple model $M$, the details of which can be found in the supplementary material.
We applied SOR to $M$ using both new weights and group lasso penalties.
We record the accuracy of the model, and the number of layer outputs removed. For simplicity, we consider an output with a weight of less than $10^{-3}$, $\beta_{ij}<10^{-3}$  to be removed. For the group lasso penalty, we consider the input channel removed if the largest weight in its group is less than  $10^{-3}$.

\begin{figure}[t!]
     \centering
     \begin{subfigure}[]{0.45\textwidth}
         \centering
         \includegraphics[width=\textwidth]{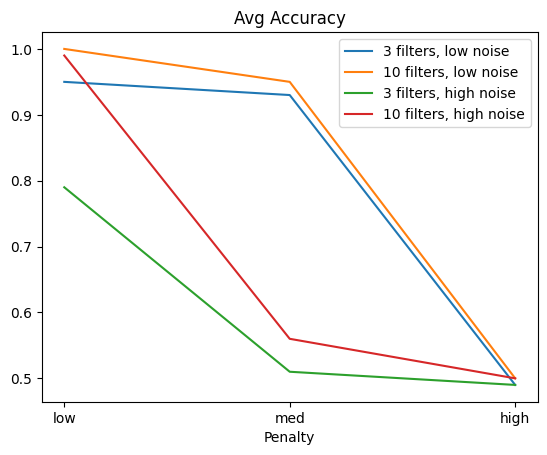}
         \caption{Model Accuracy}
         \label{nabAcc}
     \end{subfigure}
     \hfill
     \begin{subfigure}[]{0.45\textwidth}
         \centering
         \includegraphics[width=\textwidth]{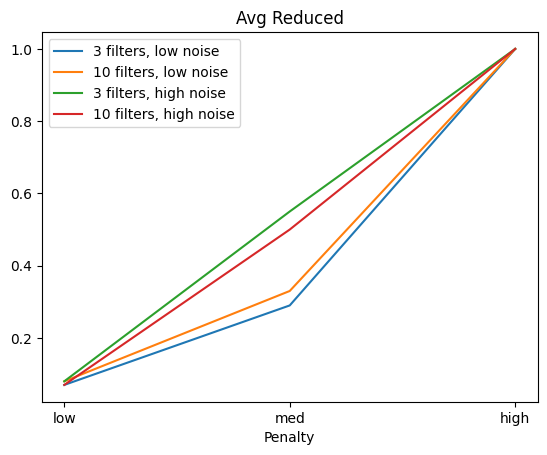}
         \caption{Amount of outputs reduced}
         \label{nabRed}
     \end{subfigure}
     \caption{Simulated results on the Noise and Box dataset with training size of 100. The accuracy chart shows the average performance of the models, while the reduced chart shows the fraction of outputs that were removed for different values of $\lambda$.}
     \label{nabResults}
\end{figure}

In Figures \ref{nabAcc} and \ref{nabRed}, we have the accuracy and model reduction achieved by SOR, for different values of $\lambda$. We can see that for low values of $\lambda$, almost nothing is removed, and high accuracy is preserved, as expected. For high values of $\lambda$ we can see that almost everything is removed, and accuracy is reduced, again, as expected. Finally, for the $\lambda$ in the middle, we can see under low noise that high accuracy is preserved, and some of the model is removed. When we increase the noise to a very high level, we can see that the accuracy is reduced. Further details including standard deviation values can be found in Tables \ref{l6} and \ref{l7first}.

These results could be explained in two ways. First, that for more difficult tasks, when optimizing for accuracy performance, not many outputs will be removed. However, for easier tasks, some outputs will be removed while maintaining performance. The second explanation is that more difficult tasks may need a lower $\lambda$ value to remove some outputs while maintaining performance.

\subsection{Few Shot Medical imaging}
Next, we tested our method on three different few-shot medical imaging classification datasets introduced to provide a benchmark for large foundation model adaptation in medical imaging applications where data is limited: a retinopathy dataset, a colon dataset, and a jaundice dataset \citep{wang2023real}.
Each training set consists of 20 percent of the data for its dataset, and they also considered 1-shot, 5-shot, and 10-shot cases where less training data is taken. For the $n$-shot training datasets, $n$ patients of each class were randomly selected from the full training set. 
In all cases the full training sets are unbalanced, but in the few shot cases they are balanced.
We ran experiments using the 5-shot and 10-shot datasets, with further experiments using the 1-shot and full datasets included in the supplementary material. We chose these two datasets because they were the few-shot cases that allowed us to have a validation set. 
We measured performance using accuracy, and each experiment was run once on a single seed.

\subsubsection{Datasets} 
The Retino dataset consists of 1392 images, with one image per patient. The task in this dataset is to classify images of retinas into one of five classes, ranging from no diabetic retinopathy to proliferative. 
The ColonPath dataset has 10009 images from 396 patients. The task for this dataset is to classify whether or not a given image has lesion tissue. 
%
The Jaundice dataset consists of head, face, and chest colour images of 745 infants. The task for this dataset is to classify the total bilirubin in the blood as high or low. 

\begin{table*}[t!]
    \centering
    \caption{Results on the 5 and 10 shot datasets. The first three columns specify the combination of dataset and model used in a particular experiment. The `Accuracy' column show the performance of our method. The `Benchmark MB' column shows the benchmark result on the dataset-model combination, with a Meta-baseline few shot learning framework. The missing values in this column are due to the benchmark paper not performing particular experiments. 
The `Benchmark Tr' column shows the benchmark result on that particular dataset-model combination, with a transfer learning framework. The missing values in this column are because the benchmark paper withheld their results for the 5-shot case due to overfitting and underfitting \citep{wang2023real}. }
    \begin{tabular}{ c c c c c c c c } 
     \hline
     Dataset & Model & Train. Size & Accuracy & Benchmark Mb & Benchmark Tr\\ 
 \hline
 
     Colon & DenseNet & 10-shot &  \textbf{87.7}  & 73.5 & 75.5\\ 
   
     Colon & DenseNet & 5-shot &  \textbf{82.7}  & 73.1 & N/A\\ 
     Colon & EfficientNet & 10-shot &  \textbf{93.9}  & N/A & 82.0\\ 
     Colon & EfficientNet & 5-shot &  \textbf{83.9}  & N/A & N/A\\ 
       \hline
   
     Retino & DenseNet & 10-shot &  \textbf{52.9}  & 46.4 & 35.9\\ 

     Retino & DenseNet & 5-shot &  40.7  & \textbf{43.2} & N/A\\ 
  
 
     Retino & EfficientNet & 10-shot &  41.2  & N/A & \textbf{57.6}\\ 

     Retino & EfficientNet & 5-shot &  \textbf{29.8}  & N/A & N/A\\ 
  \hline
 
     Jaundice & DenseNet & 10-shot  & 55.9  & 55.6 & \textbf{61.0}\\ 

     Jaundice & DenseNet & 5-shot  & 52.6  & \textbf{54.7} & N/A\\ 
 
  
     Jaundice & EfficientNet & 10-shot  & \textbf{61.8}  & N/A & 59.5\\ 
    
     Jaundice & EfficientNet & 5-shot & \textbf{59.6} &  N/A & N/A\\ 
     
    \end{tabular}
    \label{new_results_table2}
\end{table*}

\subsubsection{Application of SOR}
We ran our experiments in two stages. In the first stage we followed a similar set up to the one used in the paper that introduced the datasets \citep{wang2023real}, with a few changes made. 
As done by \cite{wang2023real}, in this stage, the convolutional base was kept frozen, and only the dense top was trained. We trained for 20 epochs, using the Adam optimizer with the default learning rate for convolutional bases Densenet121 and EfficientNetB4 respectively. The convolutional bases had weights pre-trained on ImageNet. We performed similar data augmentations, such as horizontal flips, random zoom and random rotations. We set the zoom and rotation amounts to be small, and did not test different settings of these hyperparameters. Unlike \cite{wang2023real}, we resized images to the recommended size for the particular convolutional base (e.g. 380 x 380 for EfficientNetB4, and 224 x 224 for DenseNet). Also unlike the original paper, we apply SOR by setting the frozen convolutional base to be one block, and the dense top to be a second block, including a group lasso penalty on the dense top, with the groups arranged to eliminate inputs to the layer with the penalties as described in Remark \ref{stage1}. The full training dataset that we used was the same as \cite{wang2023real}, however, for the few shot datasets, our datasets are randomly selected from the full dataset, and likely differ from those used in the original paper.

For the second stage, the first stage is used as the input. 
We allow the dense top to continue training along with the new weights $\beta$, 
allowing the model to adjust to changes in the convolutional base.
We add new weights to the outputs of blocks in the models as outlined in the examples in supplementary Section \ref{paper ex} (e.g. DenseNet blocks, and EfficientNet blocks), except we exclude the last block, as in each case it's output can be eliminated by a group lasso penalty on the dense top. We used the Adam optimizer, with the default learning rate, and ran it for 100 epochs. We did not test or optimize these hyperparameters. All other settings were kept from the first baseline. For all experiments, the $\lambda$ for both the group lasso penalty, and the regularization penalty was set to the same value. We did not test setting different values of $\lambda$. 
For each training set, we split the training data to have a validation set using about 30$\%$ of the data. Our experiments tested three values for $\lambda$, and we only used the validation sets to choose the $\lambda$ that produced the highest accuracy value.

\subsubsection{Results}

The results for our experiments on the Retino, ColonPath, and Jaundice datasets are in Table~\ref{new_results_table2}. 
%
%
Our method usually outperformed the benchmark results on the few shot datasets with DenseNet and EfficientNet. 
Our method also generally outperformed the meta-baseline few-shot methods in the benchmark paper, as well as the benchmark transfer learning results. The results for the full datasets, as well as additional experiments can be found in the supplementary materials. 
%

In the few shot settings, our method often outperformed the results from the benchmark paper when comparing the same base networks. In 8 out of 12 cases, our method produced the best results, with the benchmark transfer learning and Meta-baseline methods each producing 2 of the best results. In 4 out of 6 cases, our 5-shot results outperform the benchmark paper's 10-shot transfer learning results, and in 3 out of 6 cases our 5-shot results outperform both the 10-shot transfer learning results and the 10-shot Meta-baseline results.

\section{Conclusion}
%
%
%
In conclusion, we introduced a framework for simultaneous transfer learning and structured pruning when data is limited, that is applicable to a wide variety of models. We demonstrated a number of examples of our framework, as well as its effectiveness. 
Our method provides a way to perform transfer learning when data is very limited that is competitive with many few shot learning methods
, even when important hyperparameters such as $\lambda$ values and number of epochs are not well optimized. 
Our method also provides an avenue for models to perform data driven pruning of the whole model when data is limited. Our method generalizes well, and can be applied to a wide class of models.

Some interesting directions for future work include better optimizing $\lambda$, and optimizing the number of epochs. It would also be good to expand the models tested to include transformer based models. 
%

\subsubsection*{Acknowledgements}
The authors acknowledge the support of the Natural Sciences and Engineering Research Council of Canada (NSERC). Cette recherche a \'et\'e financ\'ee par le Conseil de recherches en sciences naturelles et en g\'enie du Canada (CRSNG),  [DGECR-2023-00311,  DGECR-2022-04531].


\bibliography{references}

\clearpage
\appendix
\thispagestyle{empty}

\onecolumn

\section{Other remarks}
\begin{remark}\label{rem2}
SOR works even when we relax the way we freeze blocks.
We first consider the case where Condition \ref{cond1} holds for the model of interest. 
When Condition \ref{cond1} holds for a given unfrozen block $f_i$, overlapping group lasso can be used to eliminate outputs from both $f_i$, and $f_{(i-1)}$ (the output and the input of $f_i$). A new weight $\beta_i$ will only be applied between two blocks that have been frozen. 
On the other hand, when Condition \ref{cond1} does not hold, then for an unfrozen block $f_i$, group lasso can only be used to eliminate outputs from block $f_{(i-1)}$. In this case, $\beta_i$ will be added to outputs of unfrozen blocks before blocks that have been frozen as well.
\end{remark}
\begin{remark}\label{rem3}
Assumption 
\ref{ass2} is also not critical for the application of SOR. When Assumption \ref{ass2} does not hold and we relax the way we freeze, our method can still be applied in the following way:
If Condition \ref{cond1} is true, then for an unfrozen block $f_i$, group lasso can only be used to eliminate outputs from block $f_{i}$. In this case, $\beta_i$ will be applied to outputs of frozen blocks.
When Condition \ref{cond1} and Assumption \ref{ass2} do not hold, group lasso cannot be used to eliminate either inputs or outputs. Therefore, $\beta_i$ should be added between all defined blocks. 
Furthermore, if either of Assumption \ref{ass2} or Condition \ref{cond1} hold, and if no blocks are frozen, then it is not appropriate to put any new weights $\beta_i$ on the outputs of any blocks.
\end{remark}

\section{Examples}\label{paper ex}
In this section, we go through some concrete examples of the application of SOR. 
\begin{remark}
    We point out that our definitions for the model and blocks do not exclude skip connections, as the input to a layer can be sent through as part of it's output. A block can be designed with some of the subfunctions $f_{ij}$ being an identity function for parts of the input.
\end{remark}

\begin{example}[DenseNet]
We now present an example where we go through each step in the method. Suppose we have a CNN with a DenseNet style base architecture \citep{huang2018denselyconnectedconvolutionalnetworks}. That is, there are DenseNet blocks that consist of a sequence of convolutional layers. Within each DenseNet block, each convolutional layer both produces an output, and sends its input through to the next layer as well. Suppose the model has three such DenseNet blocks. After the DenseNet blocks, the model has a single dense layer with an output designed for the task at hand.

The first step is to identify the $f_i$'s. In this case, we cannot have each convolutional layer be its own block $f_i$, since they send their input through unchanged as part of their output, which violates Assumption \ref{ass2}. In this case, the smallest the $f_i$'s can be (aside from the last dense layer) is one DenseNet block each. It is up the researcher whether or not they want the $f_i$'s to be larger, in which case they can combine multiple DenseNet blocks into one $f_i$. For this example, lets assume we have enough data that we are comfortable setting each DenseNet block as its own $f_i$.

The next step is to choose which $f_i$'s to freeze. For this example, lets assume that we don't have enough data to feel comfortable leaving any of the blocks unfrozen. This does not contradict our earlier assumption of having enough data for each $f_i$ to be it's own block, as there can be many more parameters inside the block than there are outputs of the block. So we now have a model split into three frozen blocks, and one unfrozen dense layer.

The next step is to identify the correct penalty types. We will put a group lasso penalty on the dense layer, grouped by input. This will work to regularize the outputs from the third block. We will put $\beta_i$'s after the first and second block, because we cannot regularize their outputs with a group lasso. We then train this new model, and prune outputs that have been eliminated, as well as any parameters that are only used to produce an eliminated output.

We can see from this example that careful planning is needed from the researcher in the first few steps to ensure that the method's foundation is implemented properly. Once properly implemented, the rest of the method is quite simple.
\end{example}
\begin{example}[EfficientNet]
    In our second example, we will show how a researcher might apply this method to an EfficientNet architecture, specifically, we will demonstrate using EfficientNet B0 \citep{chollet2015keras}. EfficientNet B0 is a model that has a sequence of seven predefined EfficientNet blocks made of various modules \citep{tan2020efficientnetrethinkingmodelscaling}. The model starts with some layers, then the sequence of EfficientNet blocks, then a few more image processing layers before a final dense layer. 

    The first step for the researcher is to identify the subfunction blocks $f_i$ to regularize. The researcher may decide that they want to regularize seven outputs from within the model, to roughly align with the seven EfficientNet blocks. In this case, the first subfunction block $f_1$, can be defined to be all initial layers in the model up to the first EfficientNet block, and the first EfficientNet block as well. Each subsequent subfunction block can be defined to be the subsequent EfficientNet block, up until the seventh subfunction block, which will include both the seventh EfficientNet block, as well as the later image processing layers in the convolutional base. The final subfunction block will be the dense layers after the convolutional base.

    For the second step, the researcher may then decide that the task specific data available is limited enough to only train the final dense layer, and a few extra parameters. In this case, the researcher could then choose to keep the last subfunction block unfrozen, and freeze all other blocks.

    The next step is to identify the appropriate penalty types. There are eight subfunction blocks total, with only the last one unfrozen. The researcher wants to regularize the outputs of the first seven subfunction blocks. Since the eighth subfunction block is unfrozen, the researcher can put a group lasso penalty on the parameters in it to regularize the outputs of the seventh subfunction block. Since the first seven subfunction blocks are frozen, the researcher cannot put a group lasso penalty on their parameters, and must intead use $\beta_i$ after the first six subfunction blocks. The new model is then trained, and the model may be pruned.
\end{example}
\begin{remark}\label{stage1}
    A common method of transfer learning in CNNs is to freeze the first $c$ layers, and to just train the last $L-c$ layers \citep{chollet2021deep,wang2023real}. One of the simpler ways to implement our method is to apply a group lasso penalty in the last $L-c$ layers to eliminate the outputs from the first $c$ layers. In this case there are only two blocks: a frozen one, and an unfrozen one that trains. As there is only one set of outputs that will be regularized, and it can be done with a group lasso penalty, we do not use any $\beta$'s in this case.
\end{remark}

\section{Data prep}
\subsection{Noise and Box generation}
We generate single channel images of size 32 x 32. Each pixel is initially set to a value drawn from a uniform distribution between zero and a given upper bound. We created two groups of datasets with two different upper bounds, 0.1 and 1.0 respectively. Then, randomly, with probability 0.5 we set a 8 x 8 square in the image to have all values be 1.0. The task for this dataset is to classify whether or not there is a 'box' (square) in the image.

\subsection{Jaundice pre-processing}
\begin{figure}
     \centering
     \begin{subfigure}[b]{0.42\textwidth}
         \includegraphics[width=\textwidth]{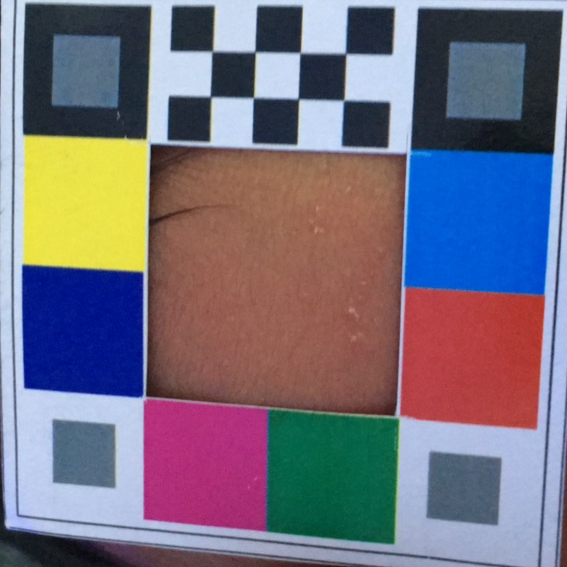}
         \caption{Before}
         \label{jaun bef}
     \end{subfigure}
     \hfill
     \begin{subfigure}[b]{0.42\textwidth}
         \includegraphics[width=\textwidth]{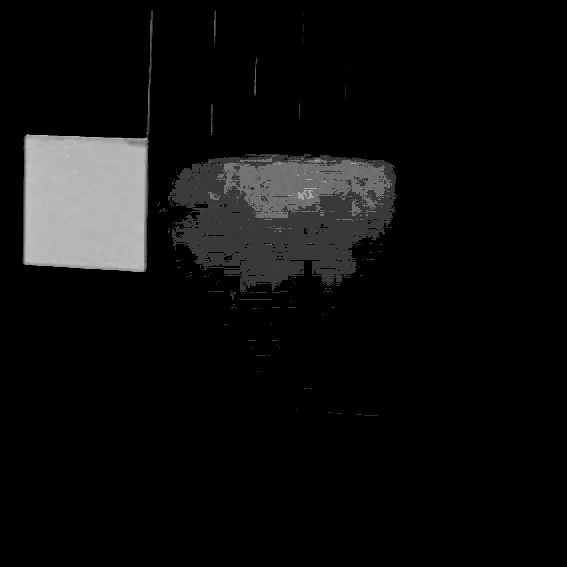}
         \caption{After}
         \label{jaun aft}
     \end{subfigure}
     \caption{Before: an image of a baby's head. After: the single channel heatmap.}
     \label{jaun}
\end{figure}

It was not clear to us if the benchmark accuracies for the Jaundice dataset were based on per image class, or per patient. To allow a single pre-trained foundation model (which takes 3 channel images as input) to make patient level predictions, we decided to pre-process the images in the Jaundice dataset to combine the three images. For each image, we first converted it from RGB to CMYK. We took the yellow channel, and made three copies. For each copy, using a different threshold, we set any pixels whose yellow value was lower than $threshold * max( cyan, magenta)$ to zero. We used thresholds of 1.1, 1.15, and 1.2, which were determined by inspecting a single image. We then averaged these three copies into a single image channel, to give a rough heatmap of areas that were more yellow. We then stacked the head, face, and chest heatmaps to make a new three channel image, with one image per patient. See example in Figure \ref{jaun}. 

\section{Supplementary experiments}
\subsection{Noise and Box details}
For each iteration, we trained a simple baseline model from scratch. The model consists of two convolutional layers, each with the same set number of filters. After the first layer, we perform max pooling, and after the second layer we perform global average pooling. The last layer of the model is a dense layer with a sigmoid activation function. The baseline was trained for 100 epochs using the Adam optimizer.

After training the baseline, we freeze the weights in the two convolutional layers. Similar to the example described in section \ref{SOR}, we add a layer $\beta_1$ after the first convolutional layer, and also add a group lasso penalty on the dense layers to perform structured output regularization. For the group lasso penalty, the weights are grouped by input channel, so that outputs from the second convolutional layer can be removed. The $\lambda$ for the group lasso is set to be 0.1 times the $\lambda$ used for the $\beta$ layer. The model is then trained for another 100 epochs, this time using the SGD optimizer, with an initial learning rate of 0.1, which gets multiplied by 0.1 every 35 epochs. We record the accuracy of the model, and the number of layer outputs removed. We consider an output with a $\beta_{ij}$ weight of less than 0.0001 to be removed. For the group lasso penalty, we consider the input channel removed if the largest weight in it's group is less than 0.0001.

\begin{table}[!]
    \centering
    \begin{tabular}{ c c c c c c } 
     \hline
     filters & $\lambda$ for $\beta$ & Average Accuracy & Accuracy STD & Average reduced & Reduced STD\\ 
     \hline
     3&0.05 & 0.9486 & 0.156 & 0.07 & 0.106\\ 
     3&0.5 & 0.933 & 0.180 & 0.288 & 0.193\\ 
     3&5 & 0.488 & 0.075 & 0.998 & 0.017\\ 
     \hline
     10&0.05 & 0.9996 & 0.003 & 0.079 & 0.055\\ 
     10&0.5 & 0.9526 & 0.136 &  0.33 & 0.086\\ 
     10&5 & 0.498 & 0.076 & 1.0 & 0.0\\ 
     \hline
    \end{tabular}
    \caption{results on the Noise and Box dataset using with training size of 100 and noise level of 0.1.}
    \label{l6}
\end{table}

\begin{table}[!]
    \centering
    \begin{tabular}{ c c c c c c } 
     \hline
     filters & $\lambda$ for $\beta$ & Average Accuracy & Accuracy STD & Average reduced & Reduced STD\\ 
     \hline
     3&0.05 & 0.786& 0.219& 0.075& 0.086\\ 
     3&0.5 & 0.510& 0.105& 0.548& 0.142\\ 
     3&5 & 0.494& 0.076& 0.998 & 0.017\\ 
     \hline
     10&0.05 & 0.990& 0.015& 0.065& 0.048\\ 
     10&0.5 & 0.563& 0.149&  0.502& 0.079\\ 
     10&5 & 0.498 & 0.076 & 1.0 & 0.0\\ 
     \hline
    \end{tabular}
    \caption{results on the Noise and Box dataset using with training size of 100 and noise level of 1.0.}
    \label{l7first}
\end{table}


\subsection{Sensitivity study on $\lambda$ for medical imaging datasets}
The first stage experiments we performed followed a similar set up to the one used in the paper that introduced the datasets \citep{wang2023real}, with a few changes made. It also provides a baseline to see what difference using $\beta_i$'s makes from doing nothing to the finished model. 
As in \citep{wang2023real}, in this baseline experiment, the convolutional base was kept frozen, and only the dense top was trained. We trained for 20 epochs, using the Adam optimizer with the default learning rate for convolutional bases Densenet121 and EfficientNetB4 respectively. The convolutional bases had weights pre-trained on ImageNet. We performed similar data augmentations, such as horizontal flips, random zoom and random rotations. We set the zoom and rotation amounts to be small, and did not test different settings of these hyperparameters. Unlike in \citep{wang2023real}, we resized images to the recommended size for the particular convolutional base (e.g. 380 x 380 for EfficientNetB4, and 224 x 224 for DenseNet). Also unlike the original paper, we apply SOR by setting the frozen convolutional base to be one block, and the dense top to be a second block, including a group lasso penalty on the dense top, with the groups arranged to eliminate inputs to the layer with the penalties as described in Remark \ref{stage1}. The full training dataset that we used was the same as in \citep{wang2023real}, however, for the 1-shot, 5-shot, and 10-shot datasets, our datasets are randomly selected from the full dataset, and likely differ from those used in the original paper.

For the next experiments, the model given is the first stage baseline as the input trained model. We allow the dense top to continue training along with the new weights. 
This allows the model to adjust to changes in the convolutional base.

We add new weights to the outputs of blocks in the convolutional base as outlined in the examples in Section \ref{paper ex} (e.g. DenseNet blocks, and EfficientNet blocks), except we exclude the last block, as in each case it's output can be eliminated by a group lasso penalty on the dense top. We used the Adam optimizer, with the default learning rate, and ran it for 100 epochs. All other settings were kept from the first baseline. For all experiments, the $\lambda$ for both the group lasso penalty, and the regularization penalty was set to the same value.

The results for our experiments on the Retino, ColonPath, and Jaundice datasets are in tables \ref{l2_2}, \ref{l3_2}, \ref{l4_3}, \ref{l7}, \ref{l8}, and \ref{l13}.  In each of these tables, the '1st stage' column represents the performance of the base model we trained and used as the input to the 2nd stage experiment, while generally following the steps outlined in the benchmark transfer learning \citep{wang2023real}. The 'BM conv. best' column represents the best reported results on the dataset, using the same particular convolutional base, and include both results from transfer learning, as well as few shot learning methods. The 'BM transfer best' column represents the best result on that particular dataset, using the same convolutional base, as well as transfer learning.

The baselines for the full datasets were generally significantly lower than the benchmark transfer learning results, despite following similar steps. A potential cause of this is the group lasso penalty that was present even in the baseline. Since we did not adjust the group lasso $\lambda$, it was likely too large or too small in many of these cases. We did not adjust the $\beta$ $\lambda$ either, and this is likely the cause of some of the poorer performance on the larger ColonPath datasets, and smaller Jaundice datasets. The ColonPath dataset is significantly larger than the others, and therefore likely needs a smaller $\lambda$, while the Jaundice dataset was the smallest and likely could have done better with a larger $\lambda$.

Our method generally performed much better for transfer learning on 1-shot, 5-shot, and 10-shot datasets with Densenet and EfficientNet compared to the benchmark paper. Our method also generally performed much better than the meta-baseline few-shot methods in the benchmark paper. Our method had mixed results compared to the benchmark's VPT few-shot method. 

\section{Other results on medical imaging datasets}

\begin{table*}[!]
    \centering
    \begin{tabular}{ c c c c c c c c } 
     \hline
     Dataset & Model & Train. Size & Best $\lambda$ & Accuracy & Benchmark Mb & Benchmark Tr\\ 
 \hline
     Colon & DenseNet & Full & 0.0005 & 76.2  & N/A & \textit{96.1}\\ 

     Colon & EfficientNet & Full & 0.0005 & 94.2  & N/A & \textit{97.}\\ 

       \hline
     Retino & DenseNet & Full & 0.0005 & 67.1  & N/A & \textit{69.9}\\

     Retino & EfficientNet & Full & 0.0005 & \textit{70.5}  & N/A & 69.6\\ 
 
  \hline
     Jaundice & DenseNet & Full & 0.0005 & 62.0  & N/A & \textit{74.2}\\

     Jaundice & EfficientNet & Full & 0.005 & 68.5  & N/A & \textit{75.2}\\

    \end{tabular}
    \caption{Full dataset results}
    \label{new_results_table}
\end{table*}

In a number of tables, such as table \ref{l7} , there are some seemingly odd results for our 2-stage method with large $\lambda$. The performance of our 2-stage method seems to get better as the amount of training data decreases. We tested to see if this is caused by using a constant batch size for all experiments, as this would cause the model to perform more updates per epoch when there is more data. If $\lambda$ is large enough that everything in the model gets eliminated after some number of updates, then in this scenario, having less updates can result in less of the model being eliminated. If enough of the model is preserved, it should retain some predictive performance.

Our results can be seen in Table \ref{batch_size_table}. We adjusted the batch sizes so that the steps per epoch were roughly the same. Interestingly, while this reduced the observed effect, it did not eliminate it. Even adjusted for the number of updates, our 2nd stage method achieved higher accuracy as the dataset size decreased.


\begin{table}[]
    \centering
    \begin{tabular}{ c c c c c c } 
     \hline
     Data & $\lambda$ & 1st stage & 2nd Stage & BM Conv. best & BM Transfer best\\ 
     \hline
     Full & 0.05 & 56.2 & 34.6 & 69.6 & 69.6\\ 
     10-shot & 0.05 & 29.1 & 18.9 & 57.6 & 57.6\\ 
     5-shot & 0.05 & 32.7 & 29.8 & N/A & N/A\\ 
     1-shot & 0.05 & 29.5 & 28.5 & N/A & N/A\\ 
     Full & 0.005 & 70.0 & 71.3 & 69.6 & 69.6\\ 
     10-shot & 0.005 & 31.3 & 43.5 & 57.6 & 57.6\\ 
     5-shot & 0.005 & 34.1 & 43.2 & N/A & N/A\\ 
     1-shot & 0.005 & 29.6 & 29.8 & N/A & N/A\\ 
     Full & 0.0005 & 68.3 & 70.5 & 69.6 & 69.6\\ 
     10-shot & 0.0005 & 31.3 & 41.2 & 57.6 & 57.6\\ 
     5-shot & 0.0005 & 33.5 & 41.1 & N/A & N/A\\ 
     1-shot & 0.0005 & 29.4 & 30.2 & N/A & N/A\\ 
     \hline
    \end{tabular}
    \caption{results on the Retino dataset using EfficientNetB4 base. 
    Data refers to the size of training set used. 1st stage is the performance of the first model we trained, while 2nd Stage is the performance of the second one. BM conv. best refers to the benchmark paper's best result when using the same convolutional base as our experiment. BM transfer refers to the benchmark paper's best result when using the same convolutional base and transfer learning.}
    \label{l3_2}
\end{table}

\begin{table}[]
    \centering
    \begin{tabular}{ c c c c c c } 
     \hline
     Data & $\lambda$ & 1st stage & 2nd Stage & BM Conv. best & BM Transfer best\\
     \hline
     Full & 0.05 & 67.0 & 60.1 & 69.9 & 69.9\\ 
     10-shot & 0.05 & 38.4 & 50.3 & 46.4 & 35.9\\ 
     5-shot & 0.05 & 45.8 & 40.7 & 43.2 & N/A\\ 
     1-shot & 0.05 & 20.6 & 32.6 & 30.6 & N/A\\ 
     Full & 0.005 & 65.9 & 72.2 & 69.9 & 69.9\\ 
     10-shot & 0.005 & 42.9 & 54.1 & 46.4 & 35.9\\ 
     5-shot & 0.005 & 46.6 & 42.1 & 43.2 & N/A\\ 
     1-shot & 0.005 & 15.5 & 21.8 & 30.6 & N/A\\ 
     Full & 0.0005 & 67.8 & 67.1 & 69.9 & 69.9\\ 
     10-shot & 0.0005 & 40.6 & 52.9 & 46.4 & 35.9\\ 
     5-shot & 0.0005 & 46.4 & 41.4 & 43.2 & N/A\\ 
     1-shot & 0.0005 & 13.8 & 27.5 & 30.6 & N/A\\ 
     \hline
    \end{tabular}
    \caption{results on the Retino dataset using DenseNet121 base. 
    Data refers to the size of training set used. 1st stage is the performance of the first model we trained, while 2nd Stage is the performance of the second one. BM conv. best refers to the benchmark paper's best result when using the same convolutional base as our experiment. BM transfer refers to the benchmark paper's best result when using the same convolutional base and transfer learning.}
    \label{l2_2}
\end{table}

\begin{table}[]
    \centering
    \begin{tabular}{ c c c c c c } 
     \hline
     Data & $\lambda$ & 1st stage & 2nd Stage & BM Conv. best & BM Transfer best\\ 
     \hline
     Full & 0.05 & 64.5 & 62.0 & 97.0 & 97.0\\ 
     10-shot & 0.05 & 78.8 & 62.0 & 82.0 & 82.0\\ 
     5-shot & 0.05 & 79.4 & 62.0 & N/A & N/A\\ 
     1-shot & 0.05 & 81.4 & 66.7 & N/A & N/A\\ 
     Full & 0.005 & 83.8 & 91.3 & 97.0 & 97.0\\ 
     10-shot & 0.005 & 88.1 & 90.3 & 82.0 & 82.0\\ 
     5-shot & 0.005 & 87.1 & 69.6 & N/A & N/A\\ 
     1-shot & 0.005 & 84.7 & 83.9 & N/A & N/A\\ 
     Full & 0.0005 & 86.8 & 94.2 & 97.0 & 97.0\\ 
     10-shot & 0.0005 & 89.4 & 93.9 & 82.0 & 82.0\\ 
     5-shot & 0.0005 & 87.7 & 83.9 & N/A & N/A\\ 
     1-shot & 0.0005 & 85.0 & 87.2 & N/A & N/A\\ 
     \hline
    \end{tabular}
    \caption{results on the ColonPath dataset using EfficientNetB4 base. 
    Data refers to the size of training set used. 1st stage is the performance of the first model we trained, while 2nd Stage is the performance of the second one. BM conv. best refers to the benchmark paper's best result when using the same convolutional base as our experiment. BM transfer refers to the benchmark paper's best result when using the same convolutional base and transfer learning.}
    \label{l4_3}
\end{table}
\begin{table}[]
    \centering
    \begin{tabular}{ c c c c c c } 
     \hline
     Data & $\lambda$ & 1st stage & 2nd Stage & BM Conv. best & BM Transfer best\\ 
     \hline
     Full & 0.05 & 84.5 & 62.0 & 96.1 & 96.1\\ 
     10-shot & 0.05 & 83.9 & 64.9 & 75.5 & 75.5\\ 
     5-shot & 0.05 & 84.4 & 75.7 & 73.1 & N/A\\ 
     1-shot & 0.05 & 80.5 & 78.4 & 68.2 & N/A\\ 
     Full & 0.005 & 87.7 & 71.6 & 96.1 & 96.1\\ 
     10-shot & 0.005 & 84.8 & 76.5 & 75.5 & 75.5\\ 
     5-shot & 0.005 & 83.4 & 82.7 & 73.1 & N/A\\ 
     1-shot & 0.005 & 80.8 & 75.6 & 68.2 & N/A\\ 
     Full & 0.0005 & 88.1 & 76.2 & 96.1 & 96.1\\ 
     10-shot & 0.0005 & 86.5 & 87.7 & 75.5 & 75.5\\ 
     5-shot & 0.0005 & 84.4 & 85.3 & 73.1 & N/A\\ 
     1-shot & 0.0005 & 80.9 & 78.9 & 68.2 & N/A\\ 
     \hline
    \end{tabular}
    \caption{results on the ColonPath dataset using DenseNet121 base. 
    Data refers to the size of training set used. 1st stage is the performance of the first model we trained, while 2nd Stage is the performance of the second one. BM conv. best refers to the benchmark paper's best result when using the same convolutional base as our experiment. BM transfer refers to the benchmark paper's best result when using the same convolutional base and transfer learning.}
    \label{l7}
\end{table}

\begin{table}[]
    \centering
    \begin{tabular}{ c c c c c c } 
     \hline
     Data & $\lambda$ & 1st stage & 2nd Stage & BM Conv. best & BM Transfer best\\ 
     \hline
     Full & 0.05 & 65.3 & 65.8 & 74.2 & 74.2\\ 
     10-shot & 0.05 & 54.8 & 55.9 & 61.0 & 61.0\\ 
     5-shot & 0.05 & 55.8 & 50.3 & 54.7 & N/A\\ 
     1-shot & 0.05 & 43.9 & 55.9 & 53.1 & N/A\\ 
     Full & 0.005 & 65.3 & 63.8 & 74.2 & 74.2\\ 
     10-shot & 0.005 & 57.1 & 56.6 & 61.0 & 61.0\\ 
     5-shot & 0.005 & 55.8 & 52.6 & 54.7 & N/A\\ 
     1-shot & 0.005 & 43.2 & 51.6 & 53.1 & N/A\\ 
     Full & 0.0005 & 64.8 & 62.0 & 74.2 & 74.2\\ 
     10-shot & 0.0005 & 57.3 & 53.3 & 61.0 & 61.0\\ 
     5-shot & 0.0005 & 55.6 & 50.3 & 54.7 & N/A\\ 
     1-shot & 0.0005 & 43.6 & 53.3 & 53.1 & N/A\\ 
     \hline
    \end{tabular}
    \caption{results on the Jaundice dataset using DenseNet121 base. 
    Data refers to the size of training set used. 1st stage is the performance of the first model we trained, while 2nd Stage is the performance of the second one. BM conv. best refers to the benchmark paper's best result when using the same convolutional base as our experiment. BM transfer refers to the benchmark paper's best result when using the same convolutional base and transfer learning.}
    \label{l8}
\end{table}


\begin{table}[]
    \centering
    \begin{tabular}{ c c c c c c } 
     \hline
     Data & $\lambda$ & 1st stage & 2nd Stage & BM Conv. best & BM Transfer best\\ 
     \hline
     Full & 0.05 & 57.0 & 57.0 & 75.2 & 75.2\\ 

     10-shot & 0.05 & 58.1 & 60.6 & 59.5 & 59.5\\ 

     5-shot & 0.05 & 61.6 & 58.5 & N/A & N/A\\ 

     1-shot & 0.05 & 53.4 & 57.5 & N/A & N/A\\ 
     Full & 0.005 & 65.8 & 68.5 & 75.2 & 75.2\\ 
     10-shot & 0.005 & 61.5 & 61.8 & 59.5 & 59.5\\ 
     5-shot & 0.005 & 61.1 & 59.6 & N/A & N/A\\ 
     1-shot & 0.005 & 54.1 & 58.3 & N/A & N/A\\ 
     Full & 0.0005 & 65.7 & 67.0 & 75.2 & 75.2\\ 
     10-shot & 0.0005 & 61.1 & 61.1 & 59.5 & 59.5\\ 
     5-shot & 0.0005 & 61.0 & 60.1 & N/A & N/A\\ 
     1-shot & 0.0005 & 54.1 & 57.5 & N/A & N/A\\ 
     \hline
    \end{tabular}
    \caption{results on the Jaundice dataset using EfficientNetB4 base. 
    Data refers to the size of training set used. 1st stage is the performance of the first model we trained, while 2nd Stage is the performance of the second one. BM conv. best refers to the benchmark paper's best result when using the same convolutional base as our experiment. BM transfer refers to the benchmark paper's best result when using the same convolutional base and transfer learning.}
    \label{l13}
\end{table}

\begin{table}[]
    \centering
    \begin{tabular}{ c c c c c c } 
    \hline
     Data & $\lambda$ & 1st stage & 2nd Stage & Batch size & Steps per epoch\\ 
     \hline
     Full & 0.05 & 86.6 & 62.0 & 80 & 25\\ 

     10-shot & 0.05 & 84.1 & 64.6 & 32 & 22\\ 

     5-shot & 0.05 & 80.8 & 66.9 & 16 & 25\\ 

     1-shot & 0.05 & 72.4 & 72.4 & 4 & 24\\ 
     \hline
    \end{tabular}
    \caption{Results on ColonPath using a DenseNet base with adjusted batch size.}
    \label{batch_size_table}
\end{table}

\end{document}